\documentclass[runningheads]{llncs}
\usepackage{graphicx}
\usepackage{comment}
\usepackage{amsmath,amssymb} 
\usepackage{color}
\usepackage{multirow}
\usepackage{multicol}
\usepackage{makecell}
\usepackage{wrapfig}
\usepackage{caption}
\usepackage{tabularx}
\usepackage{booktabs}
\usepackage{hyperref}
\usepackage{float}
\usepackage{subfig}
\usepackage[ruled,linesnumbered]{algorithm2e}
\SetAlCapNameFnt{\scriptsize}
\SetAlCapFnt{\scriptsize}
\renewcommand\footnotemark{}
\usepackage{amsfonts,mathrsfs,bm,xcolor}

\usepackage{array,arydshln}

\DeclareMathOperator*{\argmin}{arg\,min}

\DeclareMathOperator*{\minimize}{\text{minimize}}

\DeclareMathOperator*{\st}{\text{subject to}}
\DeclareMathAlphabet\mathbfcal{OMS}{cmsy}{b}{n}
\newcommand{\Def}[0]{\mathrel{\mathop:}=}

\newcommand{\name}{Ours\xspace}

\begin{document}
\pagestyle{headings}
\mainmatter
\def\ECCVSubNumber{1991}  

\title{An Image Enhancing Pattern-based Sparsity for Real-time Inference on Mobile Devices} 

\titlerunning{Pattern-based Sparsity for Real-time Mobile Inference}
%
\author{Xiaolong Ma\inst{1}\textsuperscript{$\dag$} \and
Wei Niu\inst{2}\textsuperscript{$\dag$} \and
Tianyun Zhang\inst{3} \and
Sijia Liu\inst{4} \and 
Sheng Lin\inst{1} \and 
Hongjia Li\inst{1} \and 
Wujie Wen\inst{5} \and 
Xiang Chen\inst{6} \and 
Jian Tang\inst{7} \and 
Kaisheng Ma\inst{8} \and 
Bin Ren\inst{2} \and \\ 
Yanzhi Wang\inst{1}
}
\authorrunning{X. Ma et al.}
%
\institute{Northeastern University, Boston MA 02115, USA \\
\email{\{ma.xiaol, yanz.wang\}@northeastern.edu} \and
College of William and Mary, \textsuperscript{3 }Syracuse University, \textsuperscript{4 }IBM Research, \textsuperscript{5 }Lehigh University, \textsuperscript{6 }George Mason University, \textsuperscript{7 }DiDi AI Labs, \textsuperscript{8 }Tsinghua University\\
\textsuperscript{$\dag$} Equal Contribution
}
\maketitle

\begin{abstract}
Weight pruning has been widely acknowledged as a straightforward and effective method to eliminate redundancy in Deep Neural Networks (DNN), thereby achieving acceleration on various platforms. However, most of the pruning techniques are essentially trade-offs between model accuracy and regularity 
which lead to impaired inference accuracy and limited on-device acceleration performance. 
To solve the problem, we introduce a new sparsity dimension, namely pattern-based sparsity that comprises pattern and connectivity sparsity, and becoming both highly accurate and hardware friendly. 
With carefully designed patterns, the proposed pruning unprecedentedly and consistently achieves accuracy enhancement and better feature extraction ability on different DNN structures and datasets, 
and our pattern-aware pruning framework also achieves pattern library extraction, pattern selection, pattern and connectivity pruning and weight training simultaneously. 
Our approach on the new pattern-based sparsity naturally fits into compiler optimization for highly efficient DNN execution on mobile platforms. To the best of our knowledge, 
it is the first time that mobile devices achieve real-time inference for the large-scale DNN models thanks to the unique spatial property of pattern-based sparsity and the help of the code generation capability of compilers.
\end{abstract}

\section{Introduction}


Weight pruning has been proven to be effective in eliminating redundancy in the original model~\cite{dai2019nest,wen2016learning,he2017channel}, therefore accelerating DNN execution on target computing platforms.
Non-structured pruning~\cite{han2015deep} achieves high accuracy, but is limited by its hardware unfriendliness~\cite{wen2016learning,he2017channel}. 
Meanwhile, structured pruning~\cite{wen2016learning} is hardware friendly but suffers from accuracy loss. 

\begin{figure}[t]
    \centering
    \includegraphics[width=0.95 \textwidth]{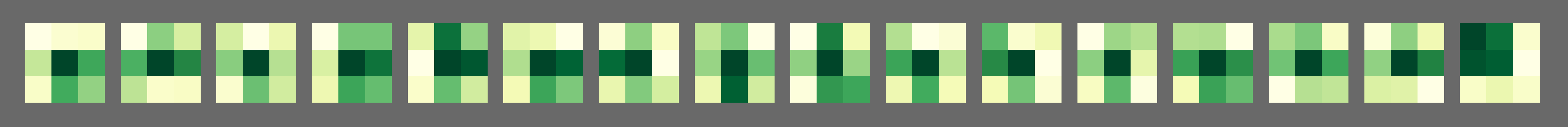}
    \caption{Heat map of randomly selected convolution kernels in the third convolutional layer of a VGG-16 on ImageNet dataset. The weight values in each kernel are normalized and darker shade represents higher absolute value.}
    \label{fig:motivation}
\end{figure}

It is imperative to seek an approach that can offer, or even go beyond, the best of both types of sparsity. 
We visualize part of the normalized heat map of a pre-trained model of VGG-16 on ImageNet in Figure~\ref{fig:motivation}, we find that (i) the effective area (i.e. weights with higher absolute values) forms some specific shapes and repeatedly appears in the model, and (ii) some of the entire convolution kernels have very small weight values and make themselves void kernels. 
Motivated by the two observations, we introduce a new sparsity dimension -- \emph{pattern-based sparsity}, which exploits both intra-convolution and inter-convolution kernel sparsities, exhibiting both high accuracy and regularity, and revealing a previously \emph{unknown} point in design space. 

In pattern-based sparsity, we call our intra-convolution kernel sparsity \emph{pattern sparsity} and inter-convolution kernel sparsity \emph{connectivity sparsity}. 
To get pattern sparsity, we prune a fixed number of weights in each convolution kernel, and the remaining weights form specific ``kernel patterns''. 
Along this line, we find that some carefully designed kernel patterns have special vision properties that potentially enhance image quality, thereby enhancing feature extraction ability of DNNs. 
For connectivity sparsity, we cut the relatively unimportant connections between certain input and output channels, which is equivalent to removal of corresponding kernels. 
At the algorithm level, we design a novel pattern-aware network pruning framework that efficiently achieves pattern pruning and connectivity pruning without degrading accuracy. 
We begin by reforming the pruning problem into an ADMM optimization problem~\cite{boyd2011distributed}, and then solve the problem iteratively using a Primal-Proximal solution which decoupling the stochastic gradient descent process with regularization, enabling a progressive and gradual process of penalizing unimportant weight groups, meaning a more accurate selection of remaining weight patterns. 
Therefore, the framework can achieve pattern library extraction, pattern assignment, unimportant connectivity removal, as well as weight training simultaneously. 
Our proposed pattern-based sparsity is mobile hardware friendly with the help of \emph{code generation} capability of compilers. More specifically, we design the \emph{filter/kernel re-ordering} technique that enables compiler optimizations that maintain instruction-level and thread-level parallelism, and achieves the maximum possible hardware acceleration. 

Our contributions of this paper are summarized as follows:
\begin{itemize}
    \item We design a set of patterns, namely pattern library, and prove the image enhancement property that is related to pattern pruning. (Section~\ref{sec:scp})
    \item We form a novel pattern-aware network pruning framework that can extract pattern library, perform pattern and connectivity pruning and weight training at the same time. (Section~\ref{sec:pattern_dist})
    \item We design the corresponding (algorithm-compiler-hardware) inference framework which fully leverages the new sparsity dimension and achieves real-time DNN execution on mobile devices. (Section~\ref{sec:connectivity_mobile_framework})
\end{itemize}
    
Section~\ref{sec:acc_results} demonstrates pattern library extraction result, pattern pruning for accuracy and image enhancement results, the overall pattern-based compression results and its acceleration results on mobile devices.

\section{Background}
\textbf{DNN model pruning techniques}
are studied in early work of \emph{non-structured pruning}~\cite{han2015deep}, in which an iterative, heuristic method is used with limited, non-uniform model compression rates. 
The irregular weight distribution causes irregular memory access and thereby execution overheads, which leads to limited acceleration performance. 
\emph{Structured pruning} is pioneered by~\cite{wen2016learning}\cite{he2017channel}, in which regular and smaller weight matrices are generated to eliminate overhead of weight indices and achieve higher acceleration in CPU/GPU executions. However, it suffers from notable accuracy drop when the pruning rate increases. 
\emph{Kernel level pruning} is studied in~\cite{chen2018sc} that the sparse complimentary kernels can save half of the weights and computations, but it is different from our approach because pattern-based sparsity is theoretically and practically improving the software and hardware performance of DNN while~\cite{chen2018sc} only focuses on parameter and computation reduction without discussing on platform acceleration.

\textbf{Mobile DNN inference frameworks} are studied, including 
TFLite~\cite{TensorFlow-Lite}, 
TVM~\cite{chen2018tvm}, 
Alibaba MNN~\cite{Ali-MNN}, 
DeepCache~\cite{xu2018deepcache} 
and DeepSense~\cite{yao2017deepsense}. 
These works do not account for model compression techniques, and the performance is far from real-time requirement (usually 30 frames/sec). There are other researches that exploit model sparsity to accelerate DNN inference 
\cite{liu2015sparse} 
\cite{parashar2017scnn}, 
but they either do not target mobile platforms (require new hardware) or trade off compression rate and accuracy, thus having different challenges than our work.

\section{Overview}\label{sec:overview}
The pattern-based sparsity should exploit the best of both non-structured and structured pruning while hiding the disadvantages. 
Given that, we propose two pattern-based pruning dimensions, \emph{pattern pruning} and \emph{connectivity pruning}.
\begin{figure}[t]
    \centering
    \includegraphics[width=0.6\textwidth]{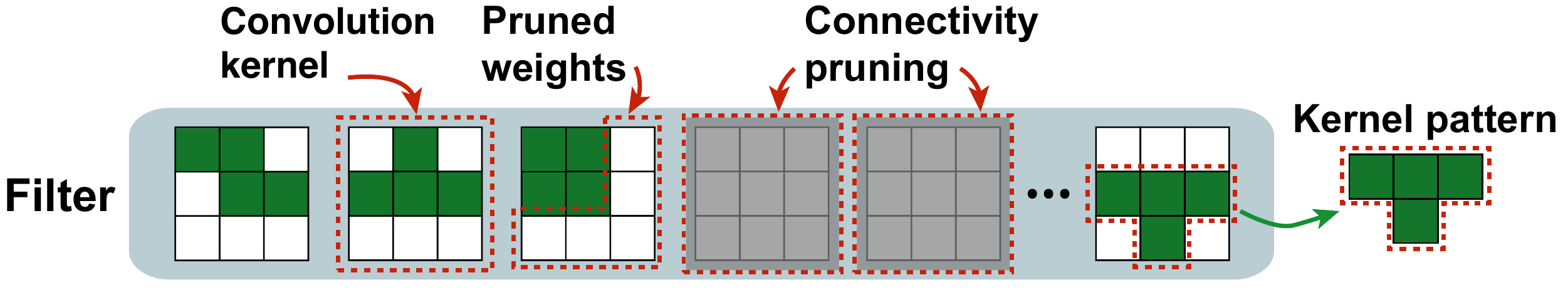}
    \caption{Illustration of pattern-based sparsity.}
    \label{fig:pattern_pruning}
\end{figure}

\textbf{Pattern pruning} is illustrated in Figure~\ref{fig:pattern_pruning}, where the white blocks denote a fixed number of pruned weights in each kernel. The remaining (four) green blocks in each kernel have arbitrary weight values, while their locations form a specific pattern. Different kernels can have different patterns, but the total number of pattern styles (i.e., the size of the pattern library) shall be limited.
We focus on 3$\times$3 kernel pattern in this work because it is widely used in various of DNN architectures. For other kernel shape (e.g., 1$\times$1 or 5$\times$5), we group 1$\times$1 kernels into 3$\times$3 then apply patterns, or use 5$\times$5 patterns directly (will not be discussed in this work due to space limit). 

\textbf{Connectivity pruning} is illustrated in Figure~\ref{fig:pattern_pruning}, with gray kernels as pruned ones. 
Connectivity pruning is a good supplement to pattern pruning, as both can be integrated in the same algorithm-level solution and compiler-assisted mobile inference framework.

\textbf{Compiler-assisted DNN inference framework} uniquely enables optimized code generation to guarantee end-to-end inference execution efficiency supporting pattern-based sparsity. As the computation paradigm of DNN is in a manner of layerwise execution, we convert a DNN model into computational graph, which is embodied by static C++ (for CPU execution) or OpenCL and CUDA (for GPU execution) codes. 
The above two pruning schemes can be naturally combined, 
which achieves high pruning (acceleration) rate while maintaining hardware friendliness. 


\section{Pattern Library -- Theory and Design}\label{sec:scp}

\subsection{A Unique Perspective on Weight Pruning}
Conventionally, weight pruning is considered as a redundant information removal technique. 
This will inevitably omit other aspects, such as the computer vision properties of pruning. In this work, we consider weight pruning as incorporating an additional convolution mask $P$ on an original kernel. $P$ has the same size as original kernels and binary-valued elements (0 and 1). From our perspective, pattern pruning is an element-wise multiplication of different $P$'s and original kernels. The set of different $P$'s is the \emph{pattern library}.

The multi-layer DNN are formed by cascading functional layers. Applying $P$ on every convolution kernel across layers is intrinsically an interpolation operation of $P$'s. Different patterns can form functional steerable filters~\cite{Freeman1991} (e.g., Gaussian blur filter, sharpen filter, edge detection filter, etc.) by interpolation, 
and this process only needs a limited number of patterns (i.e., a small pattern library). A small pattern library has two advantages, (i) at algorithm level, an appropriate number of patterns ensures the flexible search space for achieving a solution with good performance on DNN and (ii) at compiler level, fewer patterns means fewer computation paradigms after kernel reordering and grouping, which reduces thread level divergence.





\subsection{Pattern Library Design}\label{sec:scp_design}

Our designed patterns could be transformed to a series of steerable filters~\cite{Freeman1991}, which in our case, the Gaussian filter and Laplacian of Gaussian filter by interpolating patterns through DNN layers.

\textbf{Transform patterns to Gaussian filter:}
Consider a two-dimensional Gaussian filter $\mathcal G$:
\begin{equation}
\fontsize{9}{8.5}\selectfont
\setlength{\abovedisplayskip}{3pt}
\setlength{\belowdisplayskip}{3pt}
    \operatorname{\mathcal G} (x, y, \sigma)=\frac{1}{2 \pi \sigma^{2}} e^{-\frac{x^{2}+y^{2}}{2 \sigma^{2}}}\label{eq:gaussian_filter}
\end{equation}
$x$ and $y$ are input coordinates, and $\sigma^2$ is variance.

Binomial coefficients give a compact approximation of the Gaussian coefficients using only integers. To apply the Gaussian filters with $3\times3$ filter size, we utilize the following approximation. According to~\eqref{eq:gaussian_filter} and set $\sigma^2=\frac{1}{2}$, in the 1-D situation, the approximation of Gaussian filter $[1\; 2 \; 1]$ is given by the convolution of two box filters $[1\;1]$. Then we get the 2-D approximation of Gaussian filter by convolving $\left[\begin{smallmatrix}{1} & {2} & {1} \end{smallmatrix}\right]$ and $\left[\begin{smallmatrix}{1} & {2} & {1}\end{smallmatrix}\right]^T$, and the result is $\left[\begin{smallmatrix}{1} & {2} & {1} \\ {2} & {4} & {2} \\ {1} & {2} & {1}\end{smallmatrix}\right]$.

Interpolation in multi-layer DNN is proved to be convergent~\cite{powerofinterpolation}. 
We can make further approximation by interpolating patterns into convolutional layers (i.e. uniformly map patterns to each kernel). 
In continuous probability space, interpolating patterns into convolution function is a specific Probability Density Function (PDF), 
so the effect of interpolating patterns is accumulating probability expectations of interpolation into $n$ convolutional layers.
\begin{equation}
    \centering
    \includegraphics[width=0.78 \textwidth]{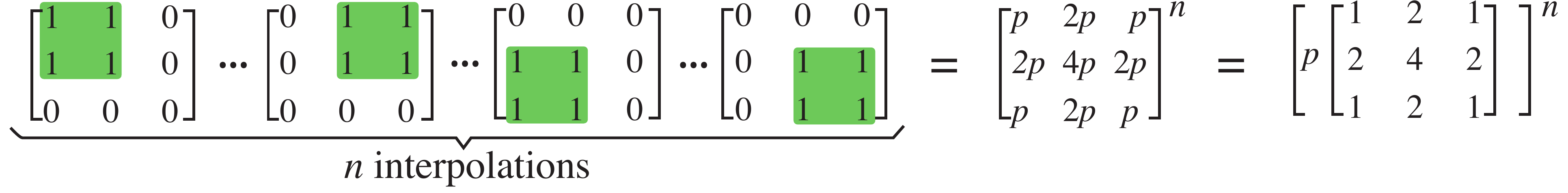}\label{eq:interpolate2}
\end{equation}

The four pattern masks $P$ shown in colored positions in \eqref{eq:interpolate2} form the Gaussian filter through interpolation. The coefficient $p$ has no effect after normalization.

\textbf{Transform patterns to Laplacian of Gaussian filter:}
The Laplacian operator is a second derivative operator.
According to the associative property, smoothing an image with Gaussian filter and then applying Laplacian operator is equivalent to convolve the image with the Laplacian of Gaussian (LoG) filter:
\begin{equation}
\fontsize{9}{8.5}\selectfont
    \nabla^{2} \mathcal G(x, y, \sigma)=\left(\frac{x^{2}+y^{2}}{\sigma^{4}}-\frac{2}{\sigma^{2}}\right) \mathcal{G}(x, y, \sigma)  
\end{equation}
LoG has elegant mathematical properties, and is valid for a variety of applications including image enhancement, edge detection, and stereo matching.

Taylor series expansion is utilized to determine the approximate values of the LoG filter with $3 \times 3$ filter size. First, we consider the 1-D situation. The Taylor series expansions of 1-D Gaussian filter $\mathcal G(x)$ are given by: 
\begin{small}
    \begin{equation}
        \mathcal G(x\!+\!\delta)\!=\!\mathcal G(x)\!+\!\delta \mathcal G^{\prime}(x)\!+\!\frac{1}{2} \delta^{2} \mathcal G^{\prime \prime}(x)\!+\!\frac{1}{3 !} \delta^{3} \mathcal G^{\prime \prime \prime}(x)\!+\!\mathcal O\left(\delta^{4}\right)\label{eq:taylor_1}
    \end{equation}
\end{small}
\begin{small}
    \begin{equation}
        \mathcal G(x\!-\!\delta)\!=\!\mathcal G(x)\!-\!\delta \mathcal G^{\prime}(x)\!+\!\frac{1}{2} \delta^{2} \mathcal G^{\prime \prime}(x)\!-\!\frac{1}{3 !} \delta^{3} \mathcal G^{\prime \prime \prime}(x)\!+\!\mathcal O\left(\delta^{4}\right)\label{eq:taylor_2}
    \end{equation}
\end{small}
By summing \eqref{eq:taylor_1} and \eqref{eq:taylor_2}, we have
\begin{equation}
\fontsize{9}{8.5}\selectfont
[\mathcal G(x-\delta)-2 \mathcal G(x)+\mathcal G(x+\delta)]/\delta^{2}\!=\!\nabla^{2} \mathcal G(x)\!+\!\mathcal O\left(\delta^{2}\right)
\label{eq:laplacian_ref}
\end{equation}
Applying central difference approximation of LoG  $\nabla^{2} \mathcal G(x)$, we derive the 1-D approximation of LoG filter as $\left[\begin{smallmatrix}{1} & {-2} & {1} \end{smallmatrix}\right]$. Then we procure the 2-D approximation of LoG filter by convolving $\left[\begin{smallmatrix}{1} & {-2} & {1} \end{smallmatrix}\right]$ and $\left[\begin{smallmatrix}{1} & {-2} & {1}\end{smallmatrix}\right]^T$, and get $\left[\begin{smallmatrix}{-1} & {2} & {-1} \\ {2} & {-4} & {2} \\ {-1} & {2} & {-1}\end{smallmatrix}\right]$ as the \emph{1st approximation}. 
According to~\eqref{eq:laplacian_ref}, we have
\begin{equation}\label{eq:plus_filter}
\fontsize{9}{8.5}\selectfont
    \nabla^{2} \mathcal G(x, y)\!=\!\left(\left[\begin{smallmatrix}{1} & {-2} & {1}\end{smallmatrix}\right]\!+\!\left[\begin{smallmatrix}{1} \\ {-2} \\ {1}\end{smallmatrix}\right]\right) * \mathcal G(x, y)
\end{equation}
Based on \eqref{eq:plus_filter}, we derive the \emph{2nd approximation} as $\left[\begin{smallmatrix}{0} & {1} & {0} \\ {1} & {-4} & {1} \\ {0} & {1} & {0}\end{smallmatrix}\right]$. 

According to the central limit theorem, the convolution of two Gaussian functions is still a Gaussian function. 
Hence, we convolve the above two approximations of LoG and then apply normalization, and get the \emph{Enhanced Laplacian of Gaussian} (ELoG) filter as 
$\left[\begin{smallmatrix}{0} & {1} & {0} \\ {1} & {8} & {1} \\ {0} & {1} & {0}\end{smallmatrix}\right]$.

Similarly, we make the further approximation by interpolating patterns into convolutional layers.

\begin{equation}
    \centering
    \includegraphics[width=0.78\textwidth]{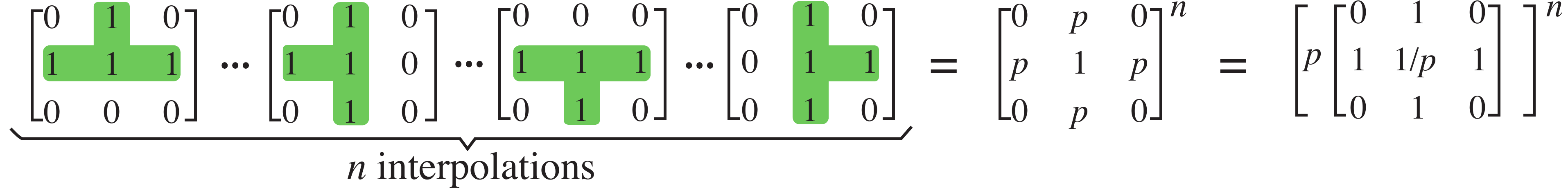}\label{eq:interpolate} 
\end{equation}
The four pattern masks $P$ shown in colored positions in \eqref{eq:interpolate} form the ELoG filter through interpolation. In order to get the best approximation to ELoG filter, 
we set $p=0.75$ and $n=8$, then the desired filter is equal to interpolating these four patterns for eight times. The coefficient $p$ has no effect after normalization.


\section{Pattern-Aware Network Pruning Framework for Pattern Library Extraction} \label{sec:pattern_dist}


In Section~\ref{sec:scp}, we have determined the (eight) patterns as our pattern library through theoretical derivation. 
However, there are still a couple of open questions. Are these theoretically derived patterns also the most desirable at algorithm level? How to select the appropriate pattern for each kernel and train corresponding (remaining) weights? To answer these questions, we propose a novel \emph{pattern-aware network pruning} framework, simultaneously achieving pattern library extraction (with predefined number of patterns in library), pattern assignment, and weight training. 

In pattern library extraction, we start from a large library comprising all possible candidate patterns. By extending ADMM~\cite{boyd2011distributed} and incorporating Primal-Proximal solution technique, we make convolution kernels dynamically ``select" the best suited patterns within the library and train the unpruned weights. Then we delete the least selected patterns in the library, thereby updating the library. The previous step is iterated on the updated library, with a single step as shown below.


\subsection{Pattern Library Extraction -- A Single Step}\label{sec:admm_formulation}

For an $N$-layer DNN of interest, let $\mathbf W$ denote the collection of weights for all $3 \times 3$ kernels, i.e., $\mathbf W = \{ \mathbf W_i \}_{i=1}^N$. The pattern of each kernel $\mathbf W_i$ is restricted to a finite pattern library $\Omega = \{ \mathbf M_1, \ldots, \mathbf M_j, \ldots, \mathbf M_{K} \}$, where $\mathbf M_j$ denotes a binary mask, and $K$ denotes the total number of possible patterns. We choose to reserve 4 non-zero entries in a kernel to match the SIMD (single-instruction multiple-data) architecture of embedded CPU/GPU processors, thereby maximizing throughput. As a result, the initial $K=\binom{9}{4}=126$, and $K$ will decrease in each step.

The purpose of each step is to select a pattern from the current library for each kernel, and train the non-zero weights.
Let $f(\mathbf W; \mathcal D)$ denote the training loss ($\mathcal D$ denotes training data), we pose the following optimization problem
\begin{align}\label{eq: prob}
    \begin{array}{cl}
\displaystyle \minimize_{\mathbf W, \mathbf z}         & f(\{ \mathbf W_i \circ (\sum_{j=1}^{K} z_j \mathbf M_j) \}_{i=1}^N; \mathcal D) \\
     \st     &  z_j \in \{ 0,1\}, \forall j, \quad \sum_{j=1}^{K} z_j = 1,
    \end{array}
\end{align}
where $z_j$ denotes the Boolean selection variable to indicate which pattern in $\Omega$ is chosen for $\mathbf W_i$.
The constraint $\sum_{j=1}^{K} z_j = 1$ indicates that only one pattern is selected, and thus
$\mathbf W_i \circ (\sum_{j=1}^{K} z_j \mathbf M_j)$  denotes the pattern-pruned kernel using one of pruning patterns. Here $\circ$ denotes element-wise product. 
In \eqref{eq: prob}, we have \textit{two} types of optimization variables: (i) $3 \times 3$ kernel weights $\mathbf W$, (ii) pattern Boolean selection variables $\mathbf z \in [0,1]^{K}$. The pattern selection scheme is co-optimized with non-zero weight training.

To solve the above problem analytically, we introduce auxiliary variables $\mathbf u $ 
together with constraints $\mathbf z = \mathbf u$. Based on that, we reformulate problem \eqref{eq: prob} as
\begin{align}\label{eq: prob1}\fontsize{9.2}{10}\selectfont
    \begin{array}{cl}
\displaystyle \minimize_{\mathbf W, \mathbf u}         & f(\{ \mathbf W_i \circ (\sum_{j=1}^{K} z_j \mathbf M_j) \}_{i=1}^N; \mathcal D) + \mathcal I(\mathbf u)\\
     \st     &  \mathbf z = \mathbf u
    \end{array}
\end{align}
where $ \mathcal I(\mathbf u)$ is the  indicator function
\begin{align}
    \mathcal I(\mathbf u) = \left \{ 
    \begin{array}{ll}
    0     &   \text{if $u_j \in [ 0,1], \forall j, \quad \sum_{j=1}^{K} u_j = 1$}\\
    \infty     & \text{otherwise}. 
    \end{array}
    \right.
\end{align}
Here we relax the binary selection variable $ z_i \in \{0,1\}$ to the (continuous) probabilistic selection variable $ u_i \in [0,1]$.

The augmented Lagrangian function of problem \eqref{eq: prob1} is given by 
\begin{align}
    \mathcal L( \mathbf W, \mathbf z, \mathbf u, \boldsymbol{\mu})& = f\big(\{ \mathbf W_i \circ (\sum\nolimits_{j=1}^{K} z_j \mathbf M_j) \}_{i=1}^N; \mathcal D\big) \\
&+ \mathcal I(\mathbf u) + \boldsymbol{\mu}^T(\mathbf z - \mathbf u) + \frac{\rho }{2} \|\mathbf z - \mathbf u \|_2^2 \nonumber
\end{align}
where $\boldsymbol{\mu} $ is Lagrangian multipliers, and $\| \cdot \|_2$ denotes the Frobenius norm. $\rho > 0$ is a given augmented penalty value, and for ease of notation we view matrices as \textit{vectors} in optimization.

ADMM is then given by the following alternating optimization process. At iteration $t$, ADMM yields
\begin{align}
&     \mathbf W^{(t)}, \mathbf z^{(t)} = \argmin_{\mathbf W, \mathbf z}  \mathcal L( \mathbf W, \mathbf z, \mathbf u^{(t-1)} , \boldsymbol{\mu}^{(t-1)})\tag{Primal} \label{eq: admm_primal} \\
& \mathbf u^{(t)} = \argmin_{\mathbf u}  \mathcal L( \mathbf W^{(t)}, \mathbf z^{(t)}, \mathbf u , \boldsymbol{\mu}^{(t-1)})  \tag{Proximal} \label{eq: admm_proximal} \\
& \boldsymbol{\mu}^{(t)} = \boldsymbol{\mu}^{(t-1)} + \rho ( \mathbf z^{(t)} - \mathbf u^{(t)} ), \label{eq:mu_update}
\end{align}
where the initial values $\mathbf u^{(0)}$ and $\boldsymbol{\mu}^{(0)}$ are given.

Problem \eqref{eq: admm_primal} can be simplified to
\begin{align}\label{eq: prob_primal_1}
    \begin{array}{ll}
\displaystyle \minimize_{\mathbf W, \mathbf z}         &  f(\{ \mathbf W_i \circ (\sum_{j=1}^{K} z_j \mathbf M_j) \}_{i=1}^N; \mathcal D)  + \frac{\rho}{2} \| \mathbf z - \mathbf a \|_2^2  
    \end{array}
\end{align}
where $\mathbf a \Def  (\mathbf u^{(t-1)} - (1/\rho) \boldsymbol{\mu}^{(t-1)})$.
In problem \eqref{eq: prob_primal_1}, the objective function is differentiable, and can thus be solved by standard DNN solvers in SGD.

Problem \eqref{eq: admm_proximal} can be equivalently decomposed over $\mathbf u$. This leads to problem
\begin{align}
   & \begin{array}{cl}
\displaystyle \minimize_{\mathbf u}         &
 \frac{\rho }{2} \| \mathbf u -  \mathbf d\|_2^2 \\
 \st & u_j \in [ 0,1], \forall j, \quad \sum_{j=1}^{K} u_j = 1,
    \end{array} \label{eq: prob_u}
\end{align}
where $\mathbf d \Def \mathbf z^{(t)} + (1/\rho) \boldsymbol{\mu}^{(t-1)}$.

Based on~\cite{parikh2014proximal}, the analytical solution to problem \eqref{eq: prob_u} is
\begin{align}
    \mathbf u^{(t)} = \left [ \mathbf d - \nu \mathbf 1 \right ]_+,
\end{align}
where $[x]_+ = x$ if $x \geq 0$ and $0$ otherwise, 
$\nu$ is the root of the equation
\begin{align}
    \mathbf 1^T \left [ \mathbf d - \nu \mathbf 1 \right ]_+ =   1.
\end{align}

Once $\mathbf W$ and $\mathbf z$ are solved, $\mathbf z$ is a continuous variable rather than a binary variable. We need an intermediate step to project continuous $\mathbf z_{\mathrm{admm}}$ to integer $\mathbf z_{\mathrm{binary}}$, yielding
\begin{align}
    \begin{array}{ll}
\displaystyle \minimize_{\mathbf z_{\mathrm{binary}}}         & \|  \mathbf z_{\mathrm{binary}} - \mathbf z_{\mathrm{admm}} \|_2^2 \\
     \st    & \mathbf 1^T \mathbf z = 1, z_i \in \{ 0, 1\}, \forall i.
    \end{array}
\end{align}
The solution is given by $[\mathbf z_{\mathrm{binary}}]_i = 1$ if $i = \mathrm{argmax}_j [\mathbf z_{\mathrm{admm}} ]_j$, and $0$ otherwise. At this point, we have simultaneously selected pattern for each kernel and trained the non-zero weights.




    

\subsection{Pattern Library Extraction -- Overall}
The overall pattern library extraction starts from $K=126$ and decreases $K$ in each step, with algorithm brief shown in Algorithm~\ref{alg:pattern_selection}. In actual implementation we set the new $K$ to be 12 in the first step as most of the patterns occur in very few times. We set the target $K$ to be either 12, 8, or 4. When the type of patterns is within this range, the overhead in code generation at compiler level can be kept small and parallelism can be maximized.

\begin{algorithm}[H]
    \scriptsize
    \textbf{Initialization: $\Omega=\{ \mathbf{M}_{1},\mathbf{M}_{2} \ldots, \mathbf {M}_{K} \}$} with $K=126$ ; \\
    \KwResult{Subsets $\Omega^{\prime}$ with $K=12,8$ or $4$;}
    \While{training neural network}{
    Update $W$ by solving \eqref{eq: admm_primal} \;
        \For{$K \gets 126$ until $K=12,8$ or $4$}{
              Solving \eqref{eq: admm_proximal} using current $\Omega$\;
              Update $\mu$ in~\eqref{eq:mu_update}\;
              Calculate pattern distribution of current $\Omega$ \; 
              Removing patterns with fewest occurrences in $\Omega$ \;
        }
    }
    
    \caption{Pattern library extraction process.}
    \label{alg:pattern_selection}
\end{algorithm}

\textbf{Total Runtime:} Despite an iterative process, the total number of epochs (and training time) can be limited. This is because except for the last step, we only need to extract a number of patterns instead of finishing the final training of non-zero weights. As a result, we can finish each step with 10\% to 20\% of the total epochs as training of the original DNN. In the last step, we need around 9 - 12 ADMM iterations, each requiring less than 20\% of the total epochs of original DNN training. So the total number of training epochs using PyTorch~\cite{paszke2019pytorch} is around 300 - 400 for the whole process, which is even lower compared with many prior art~\cite{han2015deep,molchanov2016pruning}.

\section{Connectivity Sparsity and the New Sparsity Induced Inference Framework}\label{sec:connectivity_mobile_framework}
From Section~\ref{sec:pattern_dist}, we have designed the algorithm level solution to simultaneously achieve pattern library extraction, pattern selection and weight training. 
In this section, we discuss the connectivity sparsity and how to use the same solution framework to achieve the combination of pattern sparsity and connectivity sparsity. 
We also design a compiler-assisted DNN inference framework for mobile platforms, which can fully leverages the regularity in this new sparsity type, and potentially surpasses the hardware performances with many prior works. 

\subsection{Connectivity Sparsity}
Connectivity sparsity is achieved by connectivity pruning which can be integrated in the same algorithm-level solution in Section~\ref{sec:admm_formulation} and compiler-assisted mobile inference framework. Using the same notations as in Section~\ref{sec:admm_formulation}, we define the collection of weights in $i$-th layer as $\mathbf W_i \in \mathbb{R}^{H_{i} \times W_{i} \times F_{i} \times C_{i}}$, where $H$ and $W$ denote the dimension of the convolution kernel. $F$ and $C$ denote the number of filters and channels, respectively. 
We further define critical connectivity score for each convolution kernel as 
\begin{align}
    \gamma_{i,f,c} (\mathbf W_i) = || [\mathbf W_i]_{:,:,f,c} ||_{2}
\end{align}
where $f$ and $c$ are filter and channel indices, respectively. 
The problem formulation and solution framework for achieving connectivity sparsity is similar with the ones in Section~\ref{sec:admm_formulation}. The difference is that the constraint in the framework is related to $\gamma_{i,f,c}$. 
Please note that our algorithm level solution can solve the problems of pattern pruning and connectivity pruning simultaneously or individually. 

\begin{figure}[t]
    \centering
    \includegraphics[width=0.99 \textwidth]{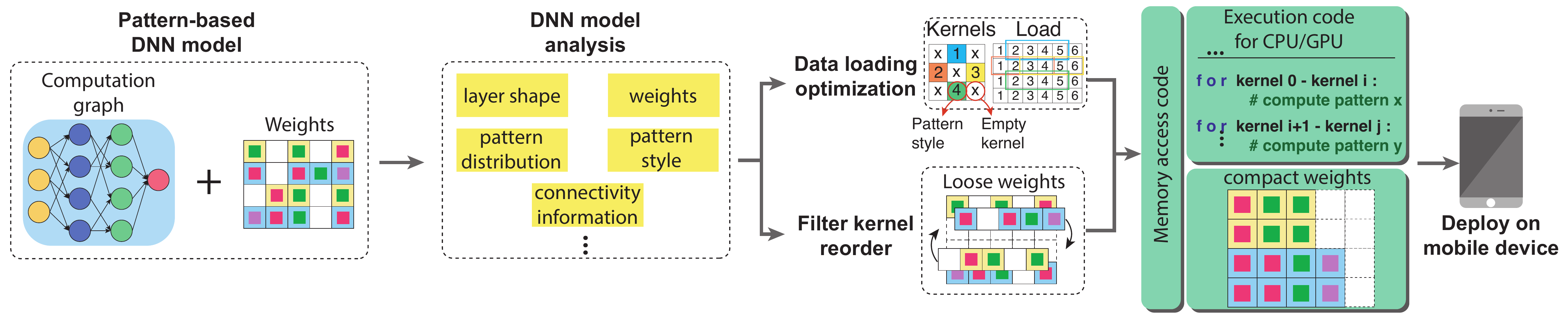}
    \caption{Overview of the compiler level DNN inference framework.}
    \label{fig:mobile}
\end{figure}

\subsection{Compiler-assisted Inference Framework for Real-time Execution}
After we obtain pattern and connectivity sparsity combined in a DNN model, we use a compiler-assisted inference framework to maximize the execution efficiency by utilizing multiple optimization techniques that are induced by pattern-based sparsity. 
The compiler optimizations showing in Figure~\ref{fig:mobile} target on DNN computation graph and memory access for on-device executions. 

\textbf{Layerwise optimization for DNN computation graph} is designed to achieve the best of instruction-level and thread-level parallelism by utilizing the unique filter/kernel re-ordering technique as Figure~\ref{fig:mobile} shows. In the weight matrix illustration, the internal squares with different colors denote different pattern styles, and empty white squares denote connectivity sparsity. By filter/kernel re-ordering, we (i) organize the filters with similar kernels together to improve {\em inter-thread} parallelism, and (ii) group kernels with identical patterns in each filter together to improve {\em intra-thread} parallelism. 
By DNN computation graph optimization, the generated execution code eliminates all of the execution branches, implying higher instruction-level parallelism; meanwhile, similar filter groups escalate execution similarity and result in a good load balance, achieving better thread-level parallelism.

\textbf{Memory access optimizations for hardware execution} address the poor memory performance due to the irregular memory access. In DNN execution, the input/output data access is associated with the non-zero elements of the weights.
Since in pattern-based sparse model, the non-zero pattern of each kernel is already known, we can generate data access code with this information for each kernel pattern and call them dynamically during DNN execution. 
With the data access code, it is possible to directly access valid input data that is associated with the non-zero elements in a pattern-based kernel. Moreover, after DNN computation graph optimization, the model weights distribution is highly compact and structured as Figure~\ref{fig:mobile} shows, which reduces the calling frequency of data access code and as a result, reduces the memory overhead.

\section{Experimental Results}\label{sec:acc_results}

In our experiment, our generated pattern-based sparse models are based on four widely used network structures, VGG-16~\cite{simonyan2014very}, ResNet-18/50~\cite{he2016deep} and MobileNet-V2~\cite{howard2017mobilenets}, and are trained on an eight NVIDIA RTX-2080Ti GPUs server using PyTorch~\cite{paszke2019pytorch}. 
We show the consistency of pattern library extraction results with the theoretically designed pattern library in Section~\ref{sec:scp_design}, and provide the accuracy improvement and image enhancement demonstrations. We also show the overall compression results of pattern-based pruning in different DNN models. 
In order to show acceleration of pattern-based sparsity on mobile devices, we compare it with three state-of-the-art DNN inference acceleration frameworks, TFLite~\cite{TensorFlow-Lite}, TVM~\cite{chen2018tvm}, and MNN~\cite{Ali-MNN}. 
Our experiments are conducted on a Samsung Galaxy S10 cell phone with the latest Qualcomm Snapdragon 855 mobile platform that consists of a Qualcomm Kryo 485 Octa-core CPU and a Qualcomm Adreno 640 GPU.

\begin{figure}[t]
    \centering
    \includegraphics[width=0.99 \textwidth]{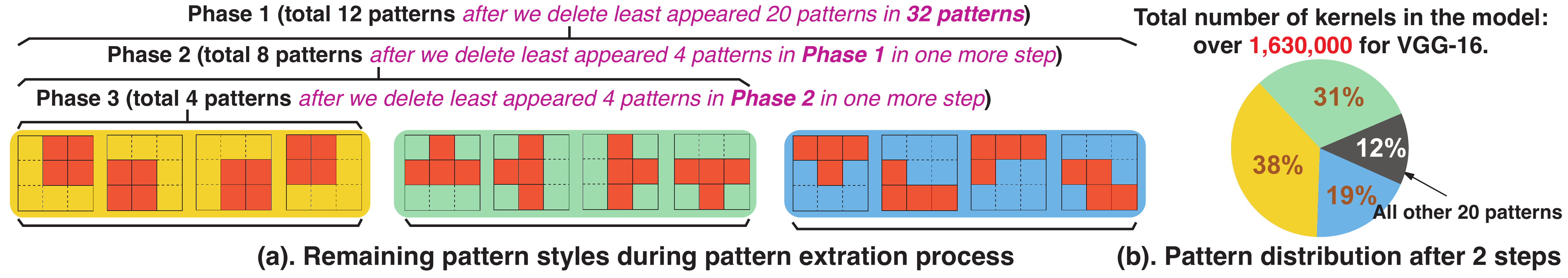}
    \caption{The pattern library extraction result. When $K=32$ after two steps, the pattern distribution is shown in \emph{(b)} with different colors representing different pattern styles in \emph{(a)}. The 20 less significant patterns only account for 13\% of the total 32 patterns, and the rest 12 patterns form the \emph{Phase 1} pattern library. If we continue the extraction step, we can get \emph{Phase 2} and \emph{Phase 3} pattern libraries as \emph{(a)} shows. }
    \label{fig:distilled_scps}
\end{figure}

\subsection{Pattern Library Extraction Result}
We use VGG-16 on ImageNet dataset to extract pattern libraries. VGG-16 has more than 1,630,000 convolution kernels. 
However, patterns can be concentrated to 12 styles in only a couple of steps. Figure~\ref{fig:distilled_scps} shows the pattern styles distribution results when $K$ decreases to 32 after two steps. We can see that most of the patterns are distributed in the top 12 styles, namely Phase 1 pattern library. If we continue to decrease $K$ to 8, the remaining 8 patterns form Phase 2 pattern library. We can notice that Phase 2 is \emph{exactly the same} with our derived pattern library in Section~\ref{sec:scp_design}. Further extraction step will give us Phase 3 pattern library, which is the top-4 pattern styles. 
Using other DNNs and datasets gives us the same extraction results, thereby we can 
conclude that the theoretically derived patterns are also the most desirable ones at algorithm level.


\subsection{Visualization Demonstration and Accuracy Analysis for Pattern Pruning}

After we obtain the extracted pattern libraries in three phases (i.e., containing 12, 8 or 4 patterns respectively), we need to validate the image enhancement effects and evaluate the accuracy of the pattern pruned DNN.

\begin{figure*}[t]
    \centering
    \includegraphics[width=0.98 \textwidth]{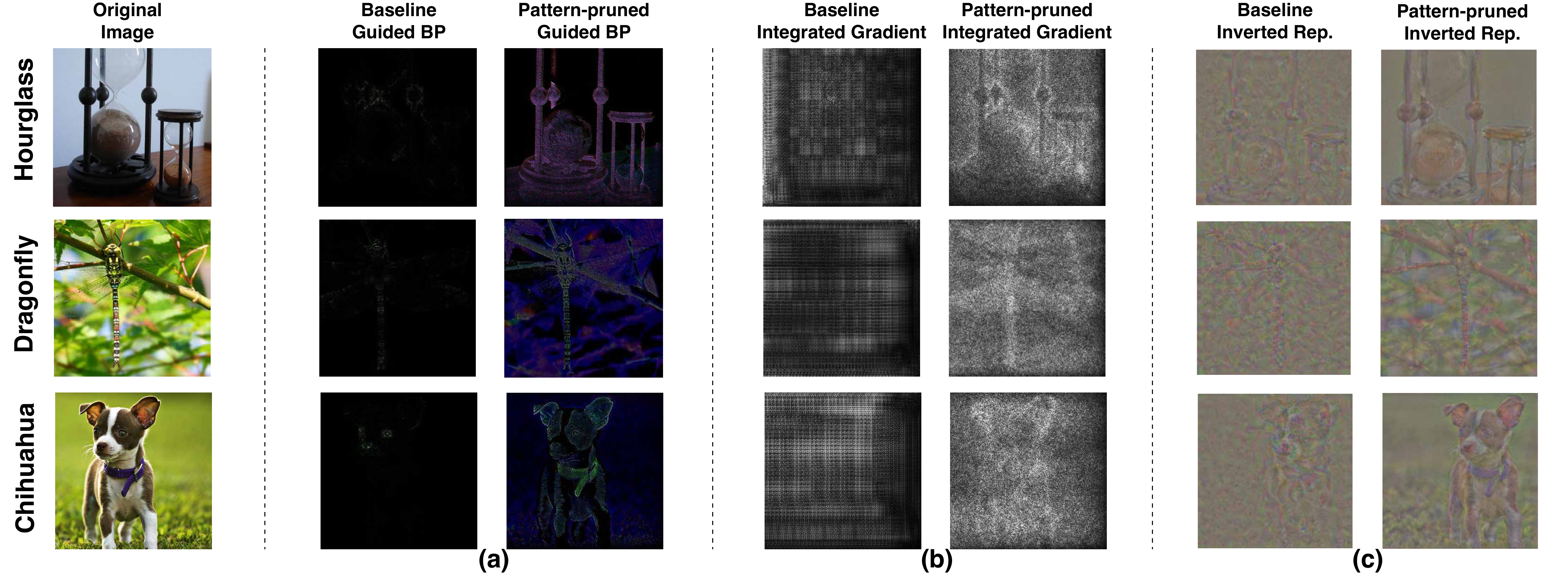}
    \caption{Visualization comparisons of three images from ImageNet dataset on original and pattern pruned VGG-16 model using (a) guided-backpropagation (BP); (b) integrated gradients and (c) inverted representation methods.}
    \label{fig:visual}
\end{figure*}

\textbf{Visualization comparisons} of applying Phase 2 pattern library to an original DNN model (\emph{pattern pruning}) are demonstrated in Figure~\ref{fig:visual}.
To ensure the fairness in comparisons, we adopt three visualization methods to eliminate the impact of causal factors. 
They are 
(a) \emph{Guided-backpropagation (BP)}~\cite{guided-backpropagation}, 
(b) \emph{Integrated gradients}~\cite{integrated_gradients}, 
and (c) \emph{Inverted representation}~\cite{inverted_representation}. 
Through different visualization techniques, we can see what a DNN has learned and how well it can preserve the photographically accurate information from an image. 

We provide strong evidence in Figure~\ref{fig:visual} that pattern pruned VGG-16 model can effectively capture more image details and less noise compared with the original VGG-16 model. We conclude that the accuracy improvement is attributed to the enhanced image processing ability of our designed pattern library.

\begin{figure}[!htbp]
    \centering
        \subfloat[]{
            \includegraphics[width=0.47 \textwidth]{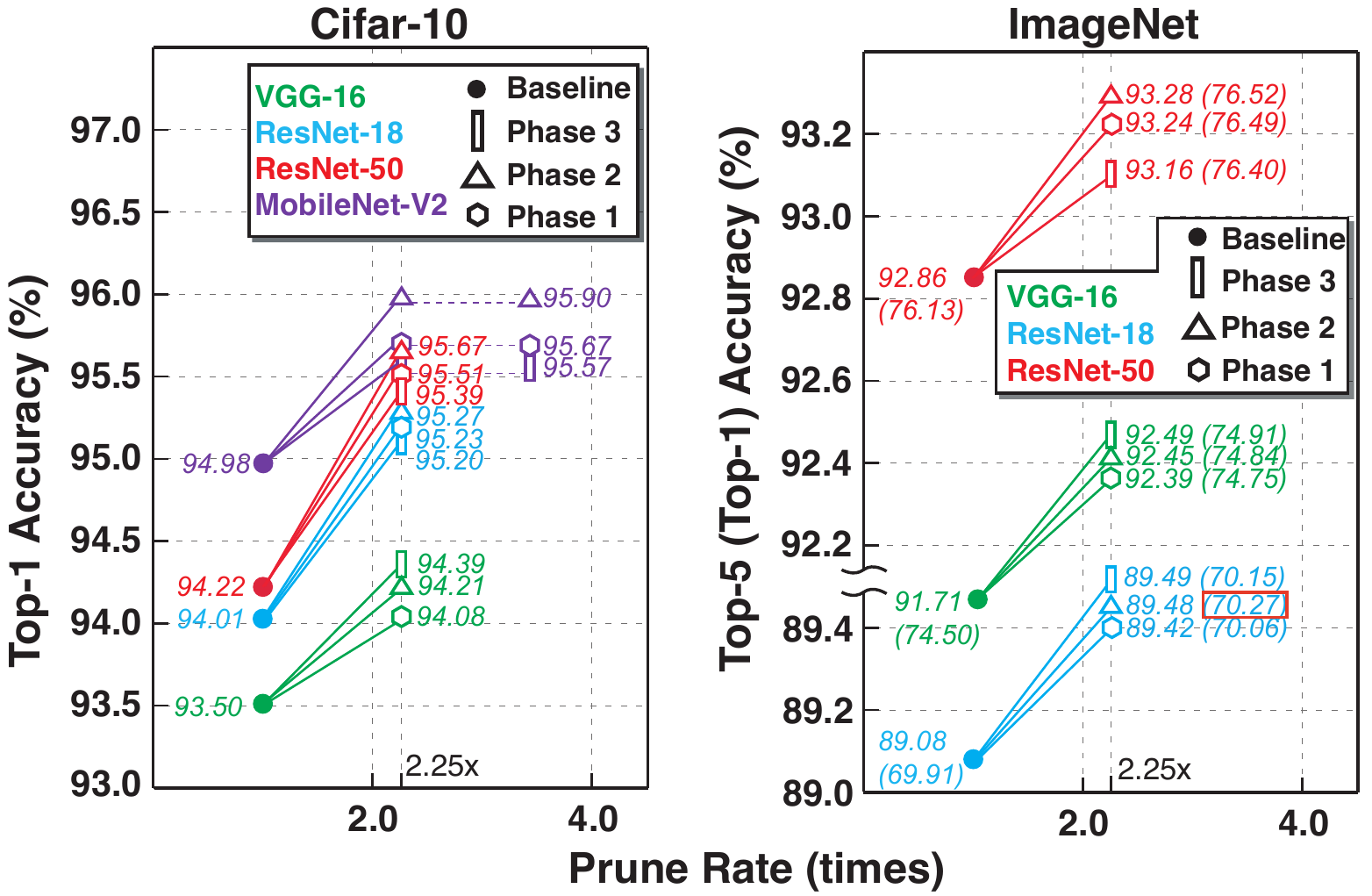}
        }
        \subfloat[]{
            \includegraphics[width=0.52 \textwidth]{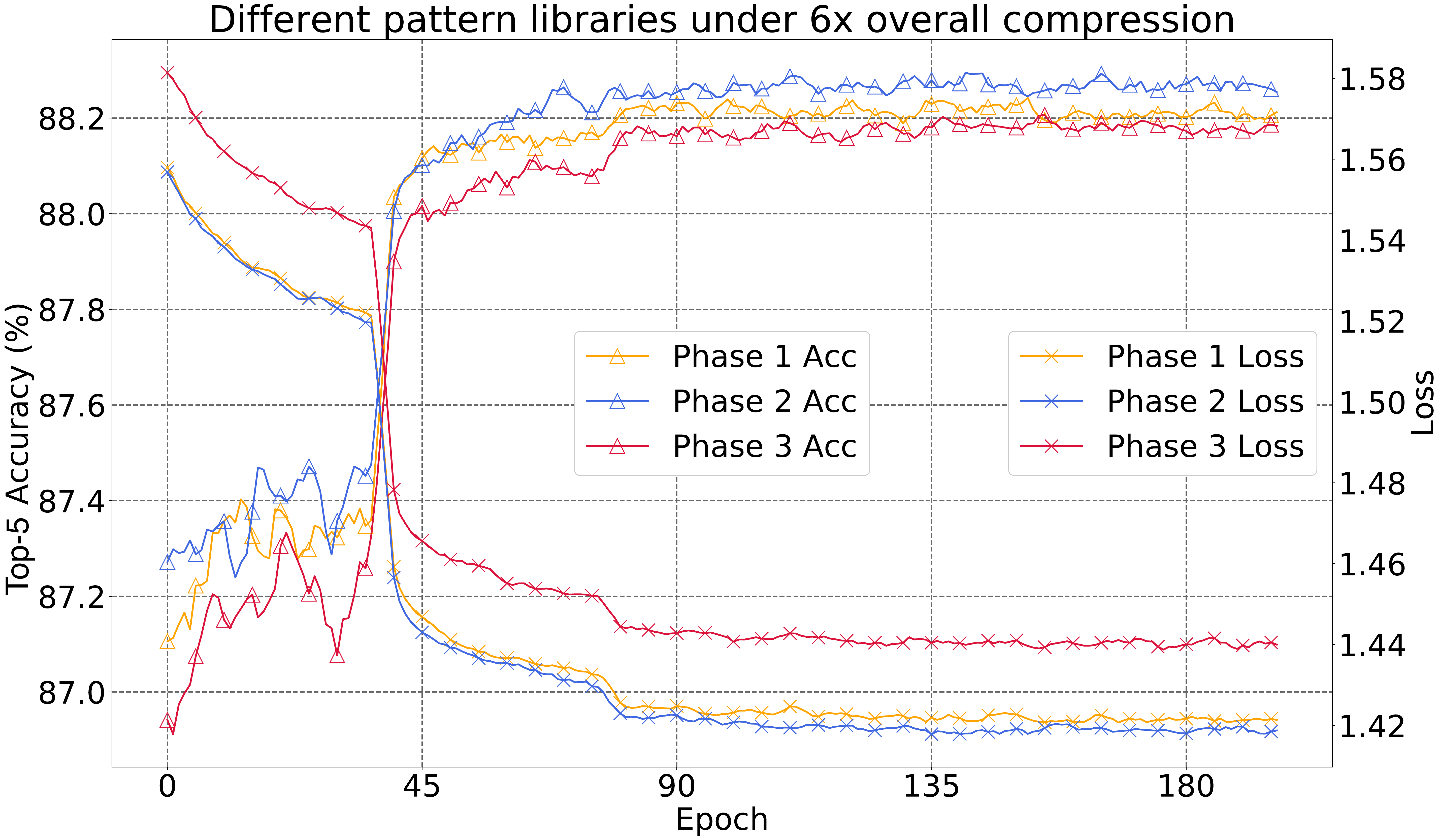}
        }
        \caption{(a) Accuracy improvement results from pattern pruning on different DNN models and datasets (CIFAR-10 \& ImageNet). (b) Overall 6$\times$ compression for ResNet-18 on ImageNet training curves for connectivity sparsity.}
    \label{fig:acc}
\end{figure}

\textbf{Accuracy evaluation} is shown in Figure~\ref{fig:acc} (a). Starting from the baseline accuracy results that are in many cases higher than prior works, we have the first conclusion that \emph{the accuracy improvements are more significant when applying the designed 8 patterns (i.e., pattern library at Phase 2) on each convolution kernel}. The accuracy improvements are consistently observed on various network structures (e.g., VGG-16, ResNet-18/50, MobileNet-V2) on CIFAR-10 and ImageNet datasets. 

\subsection{Connectivity Pruning and Overall Model Compression Results}

Combining connectivity sparsity with pattern sparsity has different DNN performances with different pattern libraries. Figure~\ref{fig:acc} (b) illustrates testing accuracies of training connectivity sparsity combined with existing pattern sparsity. From diagram, we can clearly notice that by using designed pattern library (Phase 2), we can achieve better training performance, thereby higher DNN accuracy. Similar paradigm can be observed with different compression rates and on different networks/datasets. 
Please note that pattern sparsity already reserves 2.25$\times$ compression rate, and we add different connectivity compression rates upon it to achieve the different overall compression rates. 
Table~\ref{tab:CNN} records the best final DNN accuracies and compression rates regarding their pattern styles, and are compared with several pruning methods with their sparsity types. 

\begin{table*}[!t]
\scriptsize
\centering
\caption{Pattern-based pruning results (\%) on convolution layer for CIFAR-10 and ImageNet using VGG-16, ResNet-18 and ResNet-50.}
\begin{tabular}{p{0.35cm}p{2.0cm}>{\centering}p{0.6cm}>{\centering}p{0.9cm}>{\centering}p{0.9cm}p{1.3cm}>{\centering}p{0.75cm}>{\centering}p{0.7cm}>{\centering}p{0.7cm}>{\centering}p{0.9cm}>{\centering}p{0.9cm}p{1.3cm}}
\toprule
&&\multicolumn{4}{c}{\textbf{CIFAR-10}} & \multicolumn{6}{c}{\textbf{ImageNet}} \\ \midrule
& \multirow{2}{*}{\makecell{\textbf{Pruning} \\ \textbf{Framework}}} & \multirow{2}{*}{\makecell{\textbf{Ori.} \\ \textbf{Acc.}}} & \multirow{2}{*}{\makecell{\textbf{Prune} \\ \textbf{Acc.}}} & \multirow{2}{*}{\makecell{\textbf{Comp.} \\ \textbf{Rate}}} & \multirow{2}{*}{\makecell{\textbf{Sparsity} \\ \textbf{Type}}} & \multicolumn{2}{c}{\multirow{1}{*}{\textbf{Top-1 Acc.}}} & \multicolumn{2}{c}{\multirow{1}{*}{\textbf{Top-5 Acc.}}} & \multirow{2}{*}{\makecell{\textbf{Comp.} \\ \textbf{Rate}}} & \multirow{2}{*}{\makecell{\textbf{Sparsity} \\ \textbf{Type}}} \\ \cline{7-10}
&&&&&& \textbf{Ori.} & \textbf{Prune} & \textbf{Ori.} & \textbf{Prune} && \\ \midrule
\multirow{10}{*}{\rotatebox[origin=c]{90}{ResNet-18$^{\dag}$}} & AMC~\cite{he2018amc} & 90.5 & 90.2 & 2.0$\times$ & Structured & - & - & - & - & - & - \\
& DCP~\cite{zhuang2018discrimination} & - & - & - & - & 69.6 & 64.1 & 88.9 & 85.7 & 3.3$\times$ & Structured \\
& TinyADMM~\cite{ma2019tiny} & 94.1& 93.2 & 15.1$\times$ & Structured & \emph{N/A} & \emph{N/A} & 89.1 & 88.4 & 3.3$\times$ & Structured \\
& StrADMM~\cite{zhang2018adam} & - & - & - & - & 69.6 & 68.8 & \emph{N/A} & \emph{N/A} & 3.0$\times$ & Structured \\
& SFP~\cite{he2018soft} & 92.2 & 90.8 & 1.7$\times$ & Structured & 70.3 & 67.1 & 89.6 & 87.8 & 1.7$\times$ & Structured \\
& TAS~\cite{dong2019network} & 92.8 & 92.8 & 1.8$\times$ & Structured & 70.6 & 69.1 & 89.8 & 89.2 & 1.5$\times$ & Structured \\
& FPGM~\cite{he2019filter} & 92.2 & 91.9 & 2.5$\times$ & Structured & 70.2 & 68.3 & 89.6 & 88.5 & 3.3$\times$ & Structured \\
\cdashline{2-12}[0.5pt/1pt]
& \name & 94.0 & 94.7 & 8.0$\times$ & Phase 2 & 69.9 & 69.6 & 89.1 & 89.2 & 4.0$\times$ & Phase 2 \\
& \name & 94.0 & 94.6 & 12.0$\times$ & Phase 3 & 69.9 & 68.2 & 89.1 & 88.3 & 6.0$\times$ & Phase 2 \\ 
& \name & 94.0 & 94.2 & 16.0$\times$ & Phase 2 & 69.9 & 67.1 & 89.1 & 87.7 & 8.0$\times$ & Phase 2 \\ \midrule

\multirow{10}{*}{\rotatebox[origin=c]{90}{ResNet-50$^{*}$}} & One Shot~\cite{liu2018rethinking} & 93.8  & 93.6 & 2.5$\times$ & Irregular & - & - & - & - & - & - \\
& ADMM-NN~\cite{ren2019ADMMNN} & - & - &- & - & \emph{N/A} & \emph{N/A} & \emph{N/A} & 92.3 & 7.0$\times$ & Irregular \\
& TAS~\cite{dong2019network} & 94.5 & 93.7 & 2.0$\times$ & Structured & 77.5 & 76.2 & 93.5 & 93.1 & 1.7$\times$ & Structured \\
& SFP~\cite{he2018soft} & 93.6 & 93.4 & 1.7$\times$ & Structured & 76.2 & 74.6 & 92.9 & 92.1 & 1.7$\times$ & Structured \\
& GAL~\cite{lin2019towards} & 93.3 & 90.4 & 2.9$\times$ & Structured & 76.4 & 69.3 & 92.8 & 89.1 & 2.5$\times$ & Structured \\
& FPGM~\cite{he2019filter} & 93.6 & 93.5 & 2.5$\times$ & Structured & 76.2 & 75.6 & 92.8 & 92.6 & 3.3$\times$ & Structured \\
& GBN~\cite{you2019gate} & - & - & - & - & 75.8 & 75.2 & 92.7 & 92.4 & 2.2$\times$ & Structured \\
\cdashline{2-12}[0.5pt/1pt]
& \name & 94.2 & 95.2 & 8.0$\times$ & Phase 3 & 76.1 & 75.9 & 92.9 & 92.7 & 3.9$\times$ & Phase 2 \\
& \name & 94.2 & 94.9 & 12.0$\times$ & Phase 3 & 76.1 & 75.8 & 92.9 & 92.8 & 4.9$\times$ & Phase 3 \\ 
& \name & 94.2 & 94.5 & 16.0$\times$ & Phase 3 & 76.1 & 75.6 & 92.9 & 92.6 & 5.8$\times$ & Phase 2 \\ \midrule

\multirow{8}{*}{\rotatebox[origin=c]{90}{VGG-16}} & NeST~\cite{dai2019nest} & - & - &- & - & 71.6 & 69.3 & 90.4 & 89.4 & 6.5$\times$ & Irregular \\
& ADMM-NN~\cite{ren2019ADMMNN} & - & - &- & - & 69.0 & 68.7 & 89.1 & 88.9 & 10.2$\times$ & Irregular \\
& 2PFPCE~\cite{min20182pfpce} & 92.9 & 92.8 & 4.0$\times$ & Structured & - & - & - & - & - & - \\
& DecorReg~\cite{zhu2018ijcai} & 93.5& 93.3 & 8.5$\times$ & Structured & 73.1 & 73.2 & \emph{N/A} & \emph{N/A} & 3.9$\times$ &Structured \\
& GAL~\cite{lin2019towards} & 93.9 & 90.8 & 5.6$\times$ & Structured & - & - & - & - & - & - \\
\cdashline{2-12}[0.5pt/1pt]
& \name & 93.5 & 93.4 & 8.0$\times$ & Phase 2 & 74.5 & 74.4 & 91.7 & 91.5 & 8.0$\times$ & Phase 2 \\
& \name & 93.5 & 93.3 & 11.6$\times$ & Phase 2 & 74.5 & 74.1 & 91.7 & 91.3 & 10.0$\times$ & Phase 2 \\
& \name & 93.5 & 93.2 & 19.7$\times$ & Phase 1 & 74.5 & 73.6 & 91.7 & 91.0 & 12.0$\times$ & Phase 2 \\
\bottomrule
\multicolumn{12}{l}{$\dag$ SFP, TAS, FPGM use ResNet-20 network structure on CIFAR-10 dataset.} \\
\multicolumn{12}{l}{* TAS, SFP, GAL, FPGM use ResNet-56 network structure on CIFAR-10 dataset.} \\
\end{tabular}
\label{tab:CNN}
\end{table*}

\subsection{Performance Evaluation on Mobile Platform}

In this part, we demonstrate our evaluation results on mobile device to show the real-time inference of our proposed pattern-based sparse model with the help of the compiler-assisted inference framework. 
To guarantee fairness, all frameworks are using the same pattern-based sparse model, and we also enable the fully optimized configurations of TFLite, TVM and MNN (e.g., Winograd optimization is turned on). 

\begin{figure}[tbh]
    \centering
    \includegraphics[width=0.98 \textwidth]{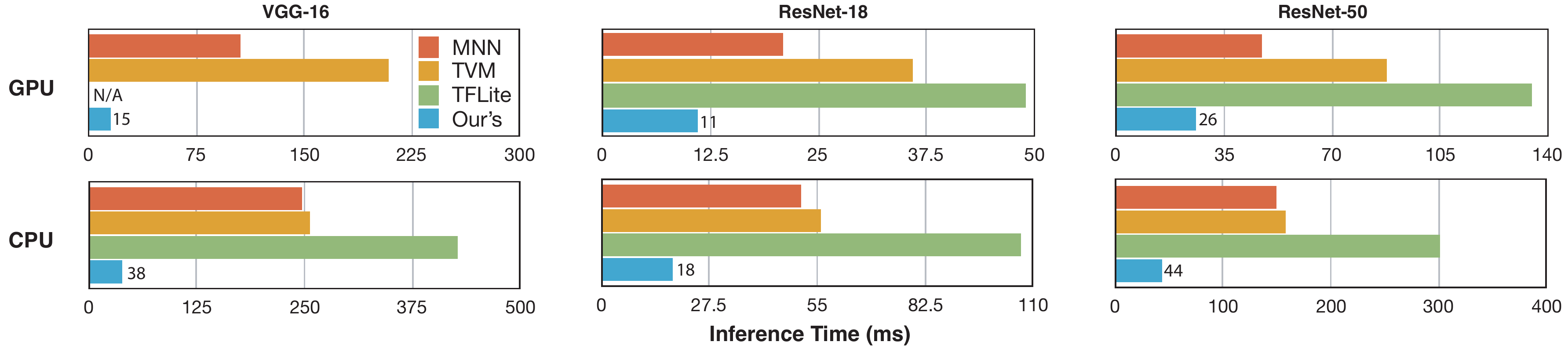}
    \caption{Inference time ($ms$) comparisons for different mobile inference frameworks using image from ImageNet dataset.}
    \label{fig:speed}
\end{figure}

\textbf{Execution time.} Figure~\ref{fig:speed} shows mobile CPU/GPU execution time of pattern-based model on different platforms. Since Phase 2 pattern library has best performance on pruning, our testing model are using Phase 2 patterns and 8$\times$ overall compression rate for ResNet-18, 5.8$\times$ for ResNet-50 and 12$\times$ for VGG-16. The inference is using images from ImageNet dataset. 
We can see our approach achieves significant acceleration on mobile device compared with other frameworks. 
Real-time execution usually requires 30 frames/sec (i.e., 33$ms$/frame). From our results, all of our DNN models on ImageNet meet or far exceed this requirement, and some of them can even accomplish real-time inference on mobile CPU. 

\section{Conclusion}
This paper proposes pattern-based sparsity, along with the highly efficient algorithm level pruning framework and the novel compiler level inference framework. 
Pattern-based sparsity inherits the flexibility from non-structured sparsity and regularity from structured sparsity, 
achieving both highly accurate/compressed model and hardware friendliness. 
Particularly, with carefully designed pattern library, pattern pruning achieves image enhancement and accuracy improvement. 
The pattern-based sparsity elicits compiler optimization, achieving real-time inference on mobile devices on various representative large-scale DNNs.

\section{Acknowledgment}
This work is supported by the National Science Foundation CCF-1919117, CCF-1937500 and CNS-1909172. 
We thank all anonymous reviewers for their feedback.

\clearpage
%
%

\end{document}